\documentclass[conference]{IEEEtran}
\IEEEoverridecommandlockouts
\usepackage{tabularx}
\usepackage{cite}
\usepackage{amsmath,amssymb,amsfonts}
\usepackage{algorithmic}
\usepackage{graphicx}
\usepackage{textcomp}
\usepackage{xcolor}
\usepackage{comment}
\usepackage{multirow} 
\usepackage{footnote}
\usepackage{url}
\usepackage{multirow}
\usepackage{makecell}
\usepackage{multirow}
\usepackage{subcaption}
\usepackage{caption}
\usepackage[title]{}
\usepackage{tabularx}
\usepackage{booktabs}
\usepackage{longtable}

\def\BibTeX{{\rm B\kern-.05em{\sc i\kern-.025em b}\kern-.08em
    T\kern-.1667em\lower.7ex\hbox{E}\kern-.125emX}}

\usepackage{times}
\usepackage{latexsym}

\usepackage{breqn}
\usepackage{amsmath}
\usepackage{url}
\usepackage[T1]{fontenc}
\usepackage[utf8]{inputenc}

\usepackage{microtype}
\usepackage{comment}

\usepackage{soul}
\usepackage{epstopdf}
\usepackage{booktabs}
\usepackage{graphicx}
\begin{document}

%\title{Exploring the Value of Pre-trained Language Models \\ for Clinical Named Entity Recognition}
\title{Investigating Large Language Models and Control Mechanisms to Improve Text Readability of Biomedical Abstracts \\
\thanks{
%We are grateful for the support from the grant “Assembling the Data Jigsaw: Powering Robust Research on the Causes, Determinants and Outcomes of MSK Disease”. The project has been funded by the Nuffield Foundation, but the views expressed are those of the authors and not necessarily the Foundation. Visit www.nuffieldfoundation.org. 
We are supported by the grant “Integrating hospital outpatient letters into the healthcare data space” (EP/V047949/1; funder: UKRI/EPSRC).
%applicable funding agency here. If none, delete this.
}
}
%Exploring the Value of Pre-trained Language Models for Clinical Named Entity Recognition

\author{
           Zihao Li $^{\dagger, 1}$ Samuel Belkadi $^{\dagger, 2}$ Nicolo Micheletti $^{\dagger, 2}$ \\ {Lifeng Han}$^{*2}$ 
           {Matthew Shardlow}$^{1}$ {Goran Nenadic}$^{2}$ \\ \vspace*{0.075cm}
            $^1$ Manchester Metropolitan University 
            $^2$ The University of Manchester\\ 
            {\tt zihao.li@stu.mmu.ac.uk }\\
            {\tt \{samuel.belkadi, nicolo.micheletti\}@student.manchester.ac.uk} \\
            {\tt \{lifeng.han,g.nenadic\}@manchester.ac.uk M.Shardlow@mmu.ac.uk } \\
            $^\dagger$ \textit{Co-first Authors} $^*$ Corresponding Author\\
}

\maketitle

\begin{abstract}
Biomedical literature often uses complex language and inaccessible professional terminologies. That is why simplification plays an important role in improving public health literacy. 
Applying Natural Language Processing (NLP) models to automate such tasks allows for quick and direct accessibility for lay readers.
In this work, we investigate the ability of state-of-the-art large language models (LLMs) on the task of biomedical abstract simplification, using the publicly available dataset for plain language adaptation of biomedical abstracts (\textbf{PLABA}). The methods applied include domain fine-tuning and prompt-based learning (PBL) on: 1) Encoder-decoder models (T5, SciFive, and BART), 2) Decoder-only GPT models (GPT-3.5 and GPT-4) from OpenAI and BioGPT, and 3) Control-token mechanisms on BART-based models. 
We used a range of automatic evaluation metrics, including BLEU, ROUGE, SARI, and BERTScore, and also conducted human evaluations. 
BART-Large with Control Token (BART-L-w-CT) mechanisms reported the highest SARI score of 46.54 and T5-base reported the highest BERTScore  72.62. 
In human evaluation, BART-L-w-CTs achieved a better simplicity score over T5-Base (2.9 vs. 2.2), while T5-Base achieved a better meaning preservation score over BART-L-w-CTs (3.1 vs. 2.6).
We also categorised the system outputs with examples, hoping this will shed some light for future research on this task.
Our codes, fine-tuned models, and data splits from the system development stage will be available at \url{https://github.com/HECTA-UoM/PLABA-MU}
\end{abstract}
\begin{IEEEkeywords}
Large Language Models, Text Simplification, Biomedical NLP, Control Mechanisms, Health Informatics
\end{IEEEkeywords}

\section{Introduction}
%\textit{what is this work about, the problem, what we tried, what findings we got. what is the structure of the rest of this paper?}

The World Health Organization (WHO) defines \textit{health literacy} as: ``the personal characteristics and social resources needed for individuals and communities to access, understand, appraise, and use information and services to make decisions about health'' \cite{world2015healthL}. 
From this, the National Health Service (NHS) of the UK emphasises two key factors for achieving better health literacy \footnote{\url{https://www.england.nhs.uk/personalisedcare/health-literacy/}}, i.e., the individual's comprehension ability and the health system itself. The ``system'' here refers to the complex network of health information and sources which promote it.  
These two factors are codependent. For instance, professionals write much healthcare information using complex language and terminologies without considering the readability of patients and the public in general.
The health system must take into account the patient's ability to achieve health literacy.
Scientific studies have reported a correlation between low health literacy, poorer health outcomes, and poorer use of health care services \cite{berkman2011low,greenhalgh2015health_Literacy}. 
Thus, Plain Language Adaptation (PLA) of scientific reports in the healthcare domain is valuable for knowledge transformation and information sharing for public patients so as to promote public health literacy \cite{10.1197/jamia.M1687_2005McCary}.
Nowadays, there have been industrial practices on such tasks, which include the publicly available plain summaries of scientific abstracts from the American College of Rheumatology (ACR) Virtual Meeting 2020 offered by the medicine company Novartis.com \footnote{\url{https://www.novartis.com/node/65241}}.

\begin{figure*}
        \centering
         \includegraphics[width=\textwidth]{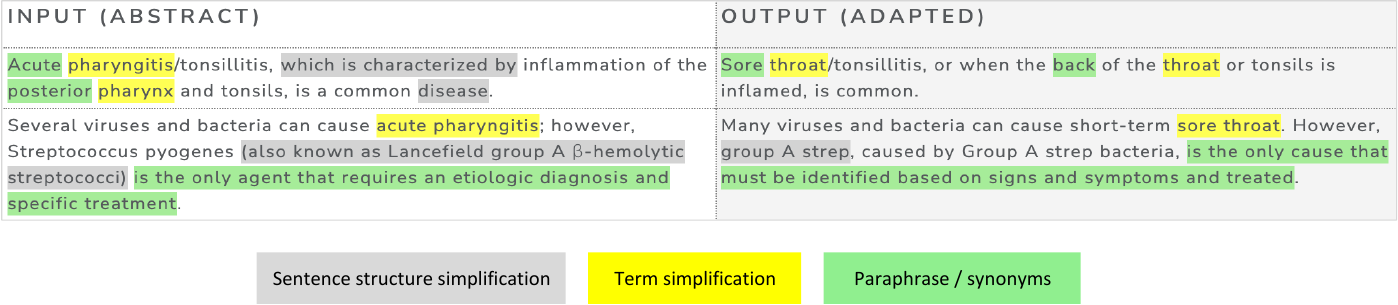}
         \caption{Examples from the PLABA dataset on Biomedical Sentences Adaptation. %\textit{to be updated with highlights using colours.} %this look good? should we also add the values? => is this from the evaluation result of 'aa' file? we can add more details on how many sentences are in each category. 
         }
          \label{fig:PLABA2023_example}
         \end{figure*}

The PLA task is related to text simplification and text summarisation, which are branches of the Natural Language Processing (NLP) field. 
This work investigates the biomedical domain PLA (BiomedPLA) using state-of-the-art large language models (LLMs) and control token methods \cite{nishihara-etal-2019-controllable_TS,martin-etal-2020-controllable_TS,agrawal-etal-2021-non_autore_TS,li-etal-2022-investigation_ControlT} that have been proven to be effective in such tasks. 
Examples of BiomedPLA can be seen in Figure \ref{fig:PLABA2023_example} from the PLABA2023 shared task\footnote{\url{https://bionlp.nlm.nih.gov/plaba2023/}}. We highlight some of the factors in colours of such tasks, including sentence simplification in grey (removing clause ``which'' in the first sentence example; separating into two sentences and removing bracket for the second example), term simplification in yellow (``pharyngitis'' and ``pharynx'' into ``throat''), paraphrasing and synonyms in green (e.g. ``acute'' into ``sore'' and ``posterior'' into ``back'' for synonyms), and summarisation (on overall text in certain situations). 
The LLMs we applied include advanced Encoder-Decoder models (T5, SciFive, BART) and Generative Pre-trained Transformers (BioGPT, ChatGPT).
The methodologies we applied include fine-tuning LLMs, prompt-based learning (PBL) on GPTs, and control token mechanisms on LLMs (BART-base and BART-large) with the efficient fine-tuning strategy.
Using the publicly available PLABA (Plain Language Adaptation of Biomedical Abstracts) data set from \cite{attal2023dataset_PLABA}, we demonstrate the capabilities of such models and carry out both quantitative and human evaluations of the model outputs.
We also discuss the interesting findings from different evaluation metrics, their inconsistency, and future perspectives on this task, based on our primary work \cite{BeeManc-plaba2023}.

\begin{comment}
    The rest of the paper is organised as below. Section \ref{sec_related} introduces related work to ours including biomedical text simplification, broader biomedical LLMs, and effective training. Section \ref{sec_method} presents different models we applied to this investigation. Section \ref{sec_evaluation} displays the experimental work and evaluation. Section \ref{sec_discuss} and \ref{sec_conclude} are the discussion and conclusion of this paper.
\end{comment}

\section{Related Work}
\label{sec_related}
%\textit{what other people did on this, related work to ours, either model or tasks.}

We first introduce recent developments in biomedical text simplification, then extend to broader biomedical LLMs related to this paper, followed by efficient training methodologies which we will apply to our work.

\subsection{Biomedical Text Simplification}

%- 'Automated lay language summarization of biomedical scientific reviews' \cite{guo2021automated_LayBio}: 
To improve the health literacy level for the general public population, \cite{guo2021automated_LayBio} developed the first lay language summarisation task using biomedical scientific reviews. The key points for this task include an explanation of context knowledge and expert language simplification. The evaluation included quality and readability using quantitative metrics.
%- Biomedical text simplification survey by \cite{ondov2022survey_biomed}:
\cite{ondov2022survey_biomed} carried out a survey, up to 2021, on biomedical text simplification methods and corpora using 45 relevant papers on this task, which data covers seven natural languages. In particular, the authors listed some published corpora on English and French languages and divided them into comparable, nonparallel, parallel, thesaurus, and pseudo-parallel. 
The quantitative evaluation metrics mentioned in these papers include SARI, BLEU, ROUGE, METEOR, and TER, among which three of them are borrowed from the machine translation (MT) field i.e., BLEU, METEOR, and TER \cite{han-etal-2021-TQA}.
Very recently, \cite{BACCO2023120874_LayLangHealth} transferred lay language style from expert physicians' notes. They developed a comparable dataset from many non-parallel corpora on plain and expert texts. The baseline model applied for training is BART, with positive outcomes. \cite{lyu2023translating_Radiology_chatGPT} did a case study using chatGPT models on translating radiology reports into plain language for patient education. Detailed prompts were discussed to mitigate the GPT models to reduce ``over-simplified'' outputs and ``neglected information''.
%'PT offers specific suggestions based on findings in the report. ChatGPT also presents some randomness in its responses with occasionally over-simplified or neglected information, which can be mitigated using a more detailed prompt.'

%Related dataset: The CDSR 

\subsection{Broader Biomedical LLMs}
Beyond text simplification tasks, there have been active developments towards biomedical domain adaptation of LLMs in recent years.
For instance, BioBERT used 4.5B words from PubMed and 13.5B words from PMC to do continuous learning based on the BERT pre-trained model. Then, it is fine-tuned in a task-specific setting on the following tasks: NER, RE, and QA \cite{10.1093/bioinformatics/btz682_BioBERT}.
In comparison, BioMedBERT \cite{chakraborty-etal-2020-biomedbert} created new data sets called BREATHE using 6 million articles containing 4 billion words from different biomedical literature, mainly from NCBI, Nature Research, Springer Nature, and CORD-19, in addition to BioASQ, BioRxiv, medRxiv, BMJ, and arXiv.
It reported better evaluation scores on QA data sets, including SQuAD and BioASQ, among other tested tasks, compared to other models.

BioMedLM 2.7B developed by Stanford Center for Research on Foundation Models (CRFM) and Generative AI company MOSAIC ML team \footnote{\url{https://www.mosaicml.com/}} formerly known as PubMedGPT 2.7B \footnote{\url{https://huggingface.co/stanford-crfm/BioMedLM}}
is trained on biomedical abstracts and papers using data from The Pile \cite{pile}. BioMedLM 2.7B claimed new state-of-the-art performance on the MedQA data set.

%'' trained exclusively on biomedical abstracts and papers from The Pile. This GPT-style model can achieve strong results on a variety of biomedical NLP tasks, including a new state-of-the-art performance of 50.3\% accuracy on the MedQA biomedical question answering charge.''

BioALBERT from \cite{naseem2021bioalbert}  is based on the ALBERT \cite{Lan2020ALBERT:} structure for training using biomedical data and reported higher evaluation scores on NER tasks of Disease, Drug, and Species, on several public data sets but much shorter training time in comparison to BioBERT. 
BioALBERT was also tested on broader BioNLP tasks using its different base and large models, including RE, Classification, Sentence Similarity, and QA by \cite{naseem2022benchmarking_bioAlbert}

Afterwards, based on the T5 model structure \cite{2020t5}, SciFive \cite{phan2021scifive} was trained on PubMed Abstract and PubMed Central (PMC) data and claimed new state-of-the-art performance on biomedical NER and RE and superior results on BioAsq QA challenge over BERT and BioBERT.
Similarly, BioBART \cite{yuan-etal-2022-biobart} was developed recently based on the new learning structure BART model \cite{lewis-etal-2020-bart}. This work will examine T5, SciFive, and BART, leaving BioBART as a future work.

\begin{comment}
    In a similar period, to explore the performances of GPT-like models in the biomedical domain, \cite{Luo2022BioGPTGP} 
pre-trained BioGPT from scratch using 15M PubMed items with titles and abstracts after filtering. BioGPT used  GPT-2 model architecture as its backbone. 
\end{comment}

Other notable related works include a) the model comparisons in biomedical domains with different tasks by \cite{lewis-etal-2020-pretrained_biome_cli,alrowili-shanker-2021-biom_LLMs, TINN2023100729_biomedLLM}  on BERT, ALBERT, ELECTRA, PubMedBERT, and PubMedELECTRA;
%2021 model: 'BioM-Transformers: Building Large Biomedical Language Models with BERT, ALBERT and ELECTRA'
%compare: 
%2020: 'Pretrained Language Models for Biomedical and Clinical Tasks: Understanding and Extending the State-of-the-Art'
b) task- and domain-specific applications on QA by \cite{alrowili2021large_biomeQA}, on Medicines by \cite{10.1001/jama.2023.14217_LLMinMed2023}, on radiology (RadBERT) by \cite{yan2022radbert}, concept normalisation by \cite{lin2022enhancing_biomeConceptNorm}, abstract generation by \cite{sybrandt2021cbag_biomedAbsG};
c) language specific models such as in French \cite{berhe-etal-2023-alibert_french} and Turkish \cite{turkmen-etal-2023-harnessing_TurkRadBERT}; and d) survey work by \cite{10.1145/3611651_plm_biome_survey}.
%lang specific French: 'Aman Berhe, Guillaume Draznieks, Vincent Martenot, Valentin Masdeu, Lucas Davy, et al.. ALIBERT: A PRETRAINED LANGUAGE MODEL FOR FRENCH BIOMEDICAL TEXT. 2023. ffhal03911564v2f'
%TurkRadBERT 

%2023: 'Pre-trained Language Models allowing Biomedical Domain: A Systematic Survey'-> survey

%task: 2021 'CBAG: Conditional biomedical abstract generation'
%2022 'RadBERT: Adapting Transformer-based Language Models to Radiology'-> domain-specific 
%and %2021 'Large Biomedical Question Answering Models with ALBERT and ELECTRA'
%Discussion on using LLMs for medicines \cite{10.1001/jama.2023.14217_LLMinMed2023} and 
%2023: 'Fine-tuning large neural language models for biomedical natural language processing for PubMedBERT-large and PubMedELECTRA

%2022: 'Enhancing Cross-lingual Biomedical Concept Normalization Using Deep Neural Network Pretrained Language Models'-> normalisation

\subsection{Efficient Training}
Due to the computational cost of the extra large-sized PLMs, some researchers proposed efficient training, which factor we will also apply to our study. These include some previously mentioned works.

\cite{houlsby2019parameter_transferNLP} proposed Parameter-Efficient Transfer Learning for NLP tasks using their Adapter modules. In this method, the parameters of the original PLMs are fixed, and a few trainable parameters are added for each fine-tuning task, between 2-4 \% of the original parameter sizes.
Using GLUE benchmark data, they demonstrated that efficient tuning with the Adapter modules can achieve similar high performances compared to the BERT models with full fine-tuning of 100\% parameters.
ALBERT \cite{Lan2020ALBERT:} applied parameter reduction training to improve the speed of BERT model learning. The applied technique uses a factorisation of the embedding parameters, which are decomposed into smaller-sized matrices before being projected into the hidden space. 
They also designed a self-supervised loss function to model the inner-sentence coherence. This reduced the parameter sizes from 108M in the BERT base to 12M in the ALBERT base models.
Addressing similar issues, \cite{li-liang-2021-prefix-tuning} proposed \textit{Prefix-tuning} method, which modifies only 0.1\% of the full parameters to achieve comparable performances using GPT-2 and BART for table-to-text generation and summarisation tasks.
%efficient fine-tuning ,   title = "Prefix-Tuning: Optimizing Continuous Prompts for Generation"
    
%BLURB here or efficient training? \cite{10.1145/3458754BLURB2021} 'BLURB is the Biomedical Language Understanding and Reasoning Benchmark.''Biomedical Language Understanding and Reasoning Benchmark (BLURB)'

Focus on the biomedical domain, 
\cite{TINN2023100729_biomedLLM} carried out fine-tuning stability investigation using the BLURB data set (Biomedical Language Understanding and Reasoning Benchmark) from \cite{10.1145/3458754BLURB2021}. Their findings show that freezing lower-level layers of parameters can be helpful for BERT-based model training, while re-initialising the top layers is helpful for low-resource text similarity tasks.

%2021: 'Fine-Tuning Large Neural Language Models for Biomedical Natural Language Processing ' -> compare and freezing parameters, low resource testing, etc. 

Instead of using Adapter modules \cite{houlsby2019parameter_transferNLP} that require additional inference latency, \cite{hu2022lora} introduced Low-Rank Adaption (LoRA) that further reduces the size of trainable parameters by freezing the weights in PLMs and injects ``trainable rank decomposition matrices'' into every single layer of the Transformer structure for downstream tasks. 
The experiments were carried out on RoBERTa, DeBERTa, and GPTs that showed similar performances compared to the Adapter modules.
We will apply LoRA for efficient fine-tuning on T5 and BioGPT in our work.

%"On GPT-2, LoRA compares favorably to both full finetuning and other efficient tuning methods, such as adapter (Houlsby et al., 2019) and prefix tuning (Li and Liang, 2021). We evaluated on E2E NLG Challenge, DART, and WebNLG:" from \url{https://github.com/microsoft/LoRA}

\section{Methodologies %and Experimental Design
}
\label{sec_method}
%\textit{{how the task is formulated and what methods/models/algorithms are used?}}

\begin{figure*}[]
\centering
    \includegraphics[width=0.9\textwidth]{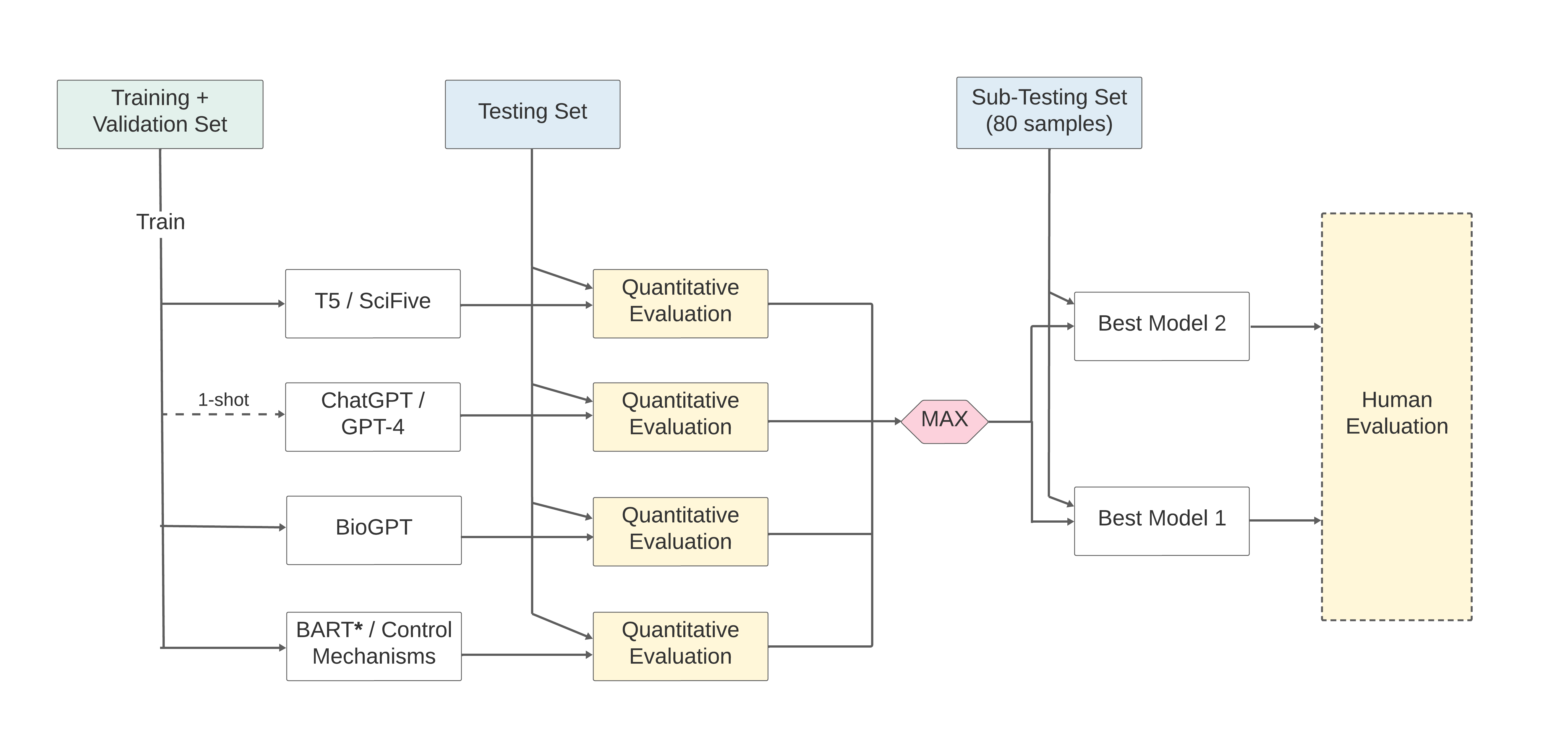}
    \caption{Model Development and Evaluation Pipeline. BART$^*$ is fine-tuned using Wikilarge data. MAX step chooses the two best-performing models according to the automatic evaluation results using SARI and BERTScore.
    %(\textbf{do we put metrics used here?}).  %Full Training set: training+validation set;  %: Fine-tuned BART / Control Mechanisms %for PLABA2023 Challenge - to be replaced with this \url{https://drive.google.com/file/d/1GKHLtuR89PBTa0HGwm70V69dAnwrv-Q-/view?usp=sharing}
    %  One-Shot ChatGPT did not use the training and validation set we extracted. One-Shot: ChatGPT/GPT4; BioGPT another chunk. BioGPT trained on train+valid.  
    }
    \label{fig:PLABA2023_pipeline}
\end{figure*}

The overall framework of our experimental design is displayed in Figure \ref{fig:PLABA2023_pipeline}. In the first step, we fine-tune selected LLMs including T5, SciFive, BioGPT, and BART, apply prompt-based learning for ChatGPTs, and optimise control mechanisms on the BART model. Then, we select the best performing two models using quantitative evaluation metrics SARI, BERTScore, BLEU and ROUGE. Finally, we chose a subset of the testing results of the two best-performing models for human evaluation.

\begin{comment}
    In this section, we first introduce the models we used, followed by LoRA efficient training, and then introduce the quantitative metrics we applied.
\end{comment}

\begin{comment} look it back later:
    there are teams/works that report high scores, but using customised data sets? -> if such data is publically available?
\end{comment}

\subsection{Models}

The models we investigated in our work include T5, SciFive, GPTs, BioGPT, BART, and Control Mechanisms; we will give more details below.
%\textit{list of models and their descriptions, and how to fine-tune them? or prompt them?}

\subsubsection{T5}
T5 \cite{2020t5} used the same Transformer structure from \cite{google2017attention} but framed the text-to-text learning tasks using the same vocabulary, sentence piece tokenisation, training, loss, and decoding. The pre-fixed tasks include summarisation, question answering, classification, and translation.
The authors used the common crawl corpus and filtered to keep only natural text and de-duplication processing. They extracted 750GB of clean English data to feed into the model for multi-task pre-training. 
Different masking strategies are integrated into the T5 model to facilitate better performances of specific fine-tuning tasks.
%The Large Language Model (LLM) T5 implemented an Encoder-Decoder Transformer architecture and was pre-trained on various sequence-to-sequence tasks. 
It has demonstrated state-of-the-art results across a wide spectrum of natural language processing tasks, showcasing its remarkable capabilities in capturing nuanced semantics and generating simplified texts while upholding high levels of accuracy. Notably, it has been successfully employed in various fields such as Clinical T5 by \cite{lehman2023clinical-t5} and \cite{lu-etal-2022-clinicalt5}. % xx-xx. \textit{give examples, preferably in the medical domain.}

\begin{comment}
    Furthermore, T5's pre-training on an extensive and diverse corpus of text data endows it with a strong foundation in understanding intricate language structures, a crucial asset for handling medical terminology. Its fine-tuning capability further enhances its adaptability, allowing us to tailor the model specifically to the nuances of the biomedical text simplification task.
These attributes make T5 a compelling candidate for this work. %our medical text simplification task.
\end{comment}

In this paper, we fine-tuned three versions of T5, namely t5-small, t5-base, and t5-large, paired with
their sentence-piece pre-trained tokenizer. Each is fine-tuned independently on the same dataset as the other models to provide comparable results. Note that we use the prompt ``summarize:” as it is the closest to our task.

\subsubsection{SciFive}
Using the framework of T5, SciFive is a Large Language Model pre-trained on the biomedical domain and has demonstrated advanced performances on multiple biomedical NLP tasks \cite{phan2021scifive}.

%While preserving the abilities of T5 in sequence-to-sequence tasks, SciFive offers a deep understanding of medical terminology, concepts, and language structures. As a result, SciFive emerges as a strong candidate for text summarisation and medical language processing tasks, offering the potential to generate clear and accurate simplifications of medical texts.

Similarly to our work on T5, we fine-tuned two versions of SciFive, namely SciFive-base and SciFive-large, paired with their pre-trained tokenizer. Each is fine-tuned independently on the same dataset as the other models to provide comparable results. We again use the prompt ``summarize:” for task-relation purposes.

\subsubsection{OpenAI's GPTs}
Given the remarkable performance demonstrated by OpenAI's GPT models in text simplification\cite{jeblick2022chatgpt}, we decided to apply simplifications using GPT-3.5-turbo and GPT-4 via its API\footnote{\url{https://openai.com/blog/openai-api}}, both models were acc.
\begin{comment}
    Example prompts we used can be found in Appendix \ref{appendix:gptprompt}.

\end{comment}

\subsubsection{BioGPT}

BioGPT \cite{Luo2022BioGPTGP} is an advanced language model specifically designed for medical text generation. BioGPT is built upon GPT-3 but is specifically trained to understand medical language, terminology, and concepts. BioGPT follows the Transformer language model backbone and is pre-trained on 15 million PubMed abstracts. It has demonstrated a high level of accuracy and has great potential for applications in medicine.
%While One-Shot prompt-based learning is applied to ChatGPT and GPT-4 without additional training using PLABA data we extracted, 
BioGPT is fine-tuned on the training and validation set, as with other encoder-decoder models (Figure \ref{fig:PLABA2023_pipeline}).

\subsubsection{BART}

Like the default transformer structure, BART \cite{lewis-etal-2020-bart} aims to address the issues in BERT and GPT models by integrating their structures with a bi-directional encoder and an autoregressive decoder. In addition to the single token masking strategy applied in BERT, BART provides various masking strategies, including deletion, span masking, permutation, and rotation of sentences. Compared to GPT models, BART provides both leftward and rightward context in the encoders.

\subsubsection{Controllable Mechanisms}
We applied the modified control token strategy in \cite{li-etal-2022-investigation_ControlT} for both BART-base and BART-large models. The training includes 2 stages, leveraging both Wikilarge training set \cite{zhang2017sentence} and our split of training set from PLABA \cite{attal_dataset_2023}.

The four attributes for control tokens (CTs) are listed below: 

\begin{itemize}
    \item \textless DEPENDENCYTREEDEPTH\_x\textgreater \space (DTD)
    \item \textless WORDRANK\_x\textgreater \space (WR)
    \item \textless REPLACEONLYLEVENSHTEIN\_x\textgreater \space (LV)
    \item \textless LENGTHRATIO\_x\textgreater \space (LR)
\end{itemize}
%\textless DEPENDENCYTREEDEPTH\_x\textgreater (DTD), \textless WORDRANK\_x\textgreater (WR), \textless REPLACEONLYLEVENSHTEIN\_x\textgreater (LV) and \textless LENGTHRATIO\_x\textgreater (LR), 

They represent  1) the syntactic complexity, 2) the lexical complexity, 3) the inverse similarity of input and output at the letter level, and 4) the length ratio of input and output respectively.

Before training, the four CTs are calculated and prepared for the 2 stage training sets. In both stages, we pick the best model in 10 epochs based on the training loss of the validation set. We applied the best model from the first stage as the base model and fine-tuned it on our PLABA training set for 10 epochs.

After fine-tuning, the next step is to find the optimal value of CTs. Following a similar process in MUSS \cite{DBLP:journals/corr/abs-2005-00352}, we applied Nevergrad \cite{nevergrad} on the validation set to find the static optimal discrete value for DTD, WR, and LV. As for the LR, we applied the control token predictor to maximise the performance with the flexible value. The predictor is also trained on Wikilarge \cite{zhang2017sentence} to predict the potential optimal value for LR.

\begin{comment}    https://www.overleaf.com/learn/latex/Positioning_images_and_tables
\end{comment}

\subsection{LoRA and LLMs}
To evaluate bigger model architectures, we fine-tune FLAN-T5 XL \cite{https://doi.org/10.48550/arxiv.2210.11416} and BioGPT-Large, which have 3 billion and 1.5 billion parameters, respectively. FLAN-T5 XL is based on the pre-trained T5 model with instructions for better zero-shot and few-shot performance. To optimise training efficiency, and as our computational resources do not allow us to fine-tune the full version of these models, we employ the LoRA \cite{hu2022lora} technique, which allows us to freeze certain parameters, resulting in more efficient fine-tuning with minimal trade-offs.

\subsection{Metrics}
We decide to evaluate our models using four quantitative metrics, namely BLEU \cite{papineni-etal-2002-bleu}, ROUGE \cite{lin-2004-rouge}, SARI \cite{xu-etal-2016-optimizing}, and BERTScore \cite{zhang2020bertscore}, each offering unique insights into text quality. SARI and BERTScore are used from EASSE \cite{alva-manchego-etal-2019-easse} package; BLEU and ROUGE metrics are imported from the Hugging Face\footnote{\url{https://github.com/huggingface/evaluate}} implementations. 

While BLEU quantifies precision by assessing the overlap between n-grams in the generated text and references, ROUGE measures recall by determining how many correct n-grams in the references are present in the generated text. This combination makes them useful as an initial indicative evaluation for machine translation and summarisation quality.

In contrast, SARI goes beyond n-gram comparisons and evaluates fluency and adequacy in translations. It does this by considering precision (alignment with references), recall (coverage of references), and the ratio of output length to reference length. SARI's comprehensive approach extends its utility to broader evaluations of translation quality.

Finally, BERTScore delves into the semantic and contextual aspects of text quality. Using a pre-trained BERT model, it measures the similarity between word embeddings in the generated and reference texts. This metric provides insight into the semantic similarity and contextual understanding between generated and reference texts, making it akin to human evaluation. This metric does not quantify how good the simplification is but rather how much the meaning is preserved after simplification.

This comprehensive evaluation effectively addresses surface-level and semantic dimensions, resulting in a well-rounded and thorough assessment of the quality of machine-generated simplifications.

\section{Experiments and Evaluations}
\label{sec_evaluation}

%We used for the dataset provided as part of the PLABA task for model development  %2023 shared task \url{https://bionlp.nlm.nih.gov/plaba2023/}created by \cite{attal2023dataset_PLABA}.
PLABA data set is extracted from PubMed search results using 75 healthcare-related questions that MedlinePlus users asked. It includes 750 biomedical article Abstracts manually simplified into 921 adaptations with 7,643 sentence pairs in total. 
The dataset is publicly available via Zenodo \footnote{\url{https://zenodo.org/record/7429310}}. 

\subsection{Data Preprocessing and Setup}
To investigate the selected models for training and fine-tuning, we divided the PLABA data into Train, Validation, and Test sets, aiming for an 8:1:1 ratio. However, in the final implementations, we found that there are only a few 1-to-0 sentence pairs, which might cause a negative effect in training the simplification models. 
Thus we eliminated all 1-to-0 sentence pairs. In addition, to better leverage the SARI score, we picked sentences with multi-references for validation and testing purposes. 
As a result, we ended up with the following sentence pair numbers according to the source sentences (5757, 814, 814).

\begin{comment} to look back:
\textit{For comparison, shall we also run our best model re-training on all PLABA training sets instead of 80\% of it?}    
\end{comment}

% \subsection{Model Size Variations with LoRA}
% Table \ref{tab:LoRA_size_impact} presents that model size reduction using LoRA. FLAN-T5 XL significantly reduces its trainable parameters, from 2.86 billion 9.5 million parameters, by applying the LoRA efficient training technique.

\begin{comment}
    \begin{table}[htb!]
    \centering
    \begin{tabular}{cp{1.5cm}c} \hline 
        Model & full size & with LoRA \\
        FLAN-T5 XL & 2.86 B  & 9.5 M parameters \\
        BioGPT & xxx & xxx \\ \hline 
    \end{tabular}
    \caption{The Impact on Models Sizes with LoRA}
    \label{tab:LoRA_size_impact}
\end{table}
\end{comment}

\subsection{Automatic Evaluation Scores}

In this section, we list quantitative evaluation scores and some explanations for them. 
The results for T5 Small, T5 Base, T5 Large, FLAN-T5 XL with LORA, SciFive Base, SciFive Large, and BART models with CTs (BART-w-CTS) are displayed in Table \ref{table:evaluation_metrics_t5_sci5_bart_test}.
Interestingly, the fine-tuned T5 Small model obtains the highest scores in both BLEU and ROUGE metrics including ROUGE-1, ROUGE-2, and ROUGE-L. 
The fine-tuned BART Large with CTs produces the highest SARI score at 46.54; while the fine-tuned T5 Base model achieved the highest BERTScore (72.62) with a slightly lower SARI score (44.10). The fine-tuned SciFive Large achieved the highest SARI score (44.38) among T5-like models, though it is approximately 2 points lower than BART Large with CTs.

The quantitative evaluation scores of GPT-like models are presented in Table \ref{table:evaluation_metrics_gpt_test} including GPT-3.5 and GPT-4 using prompts, and fine-tuned BioGPT with LoRA.
GPT-3.5 reported relatively higher scores than GPT-4 on all lexical metrics except for SARI, and much higher score on BERTScore than GPT-4 (58.35 vs 46.99).
In comparison, BioGPT-Large with LoRA reported the lowest SARI score (18.44) and the highest BERTScore (62.9) among these three GPT-like models.
Comparing the models across Table \ref{table:evaluation_metrics_t5_sci5_bart_test} and Table \ref{table:evaluation_metrics_gpt_test}, the GPT-like models did not beat T5-Base on both SARI and BERTScore, and did not beat BART-w-CTs on SARI.

\begin{figure}[]
  \centering
  \begin{subfigure}{0.49\textwidth}
    \includegraphics[width=\linewidth]{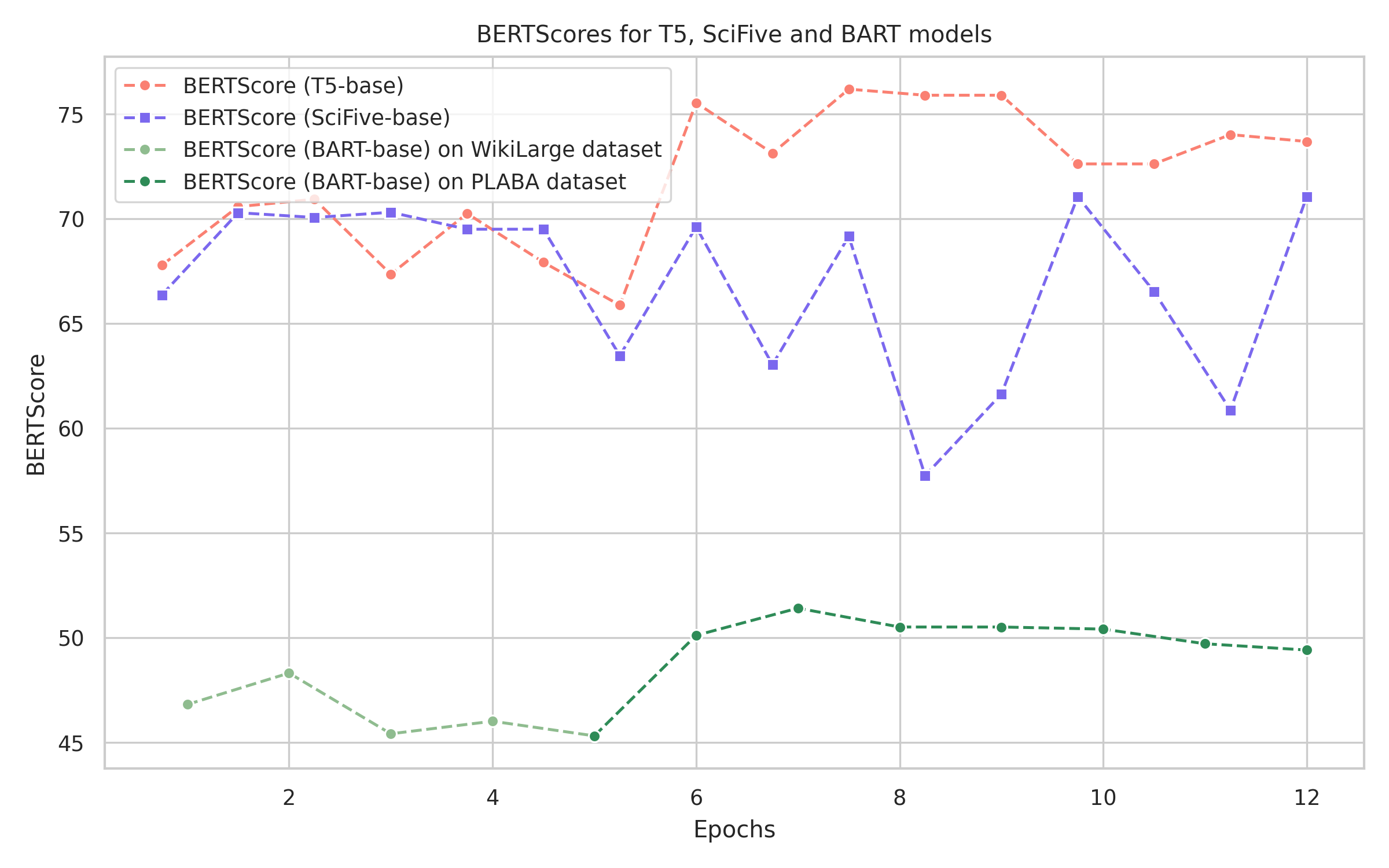}
  \end{subfigure}
  \hfill
  \begin{subfigure}{0.49\textwidth}
    \includegraphics[width=\linewidth]{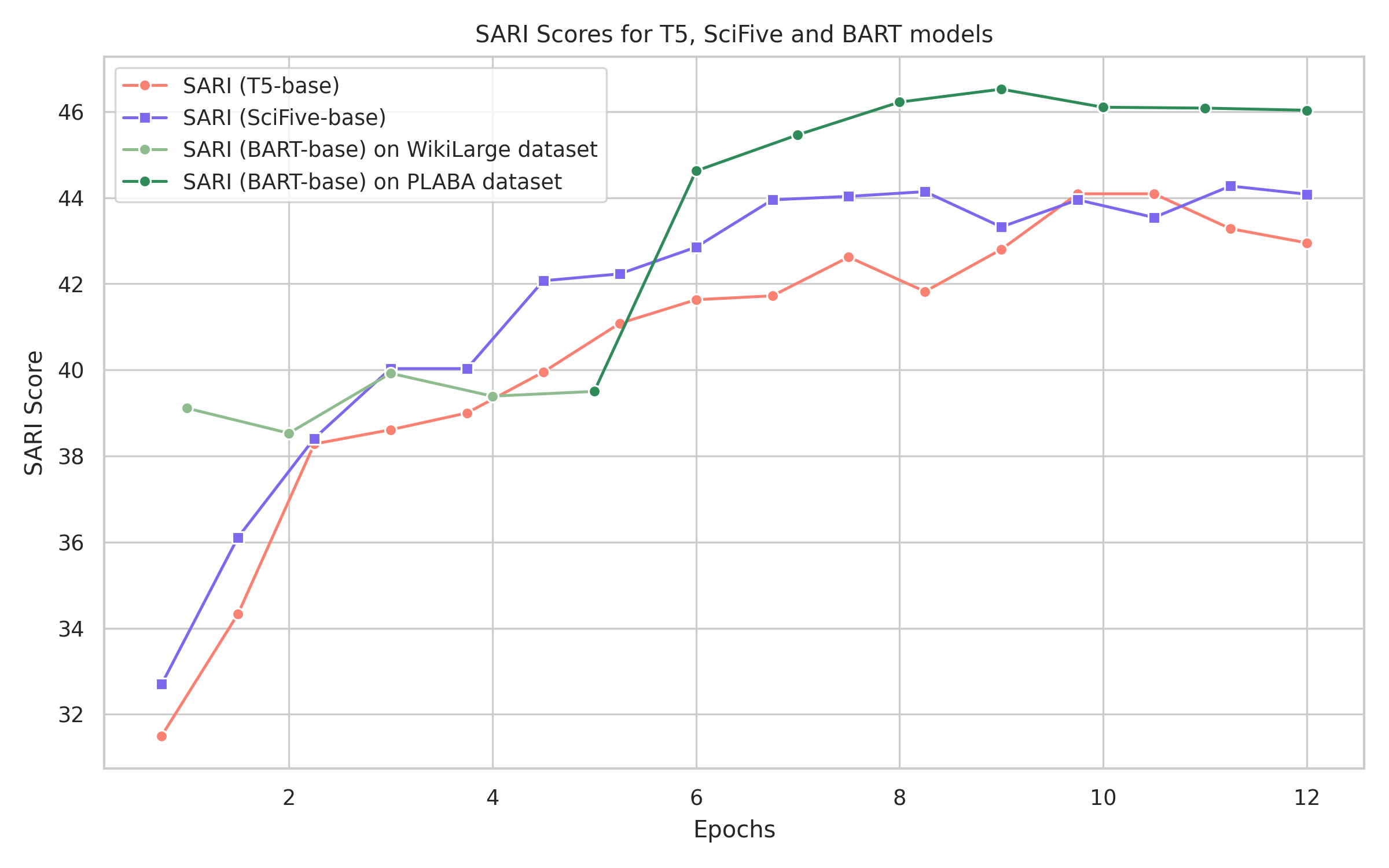}
  \end{subfigure}
  \caption{ Evaluation Scores of T5, SciFive and BART Models on the Extracted Testing Set}
  \label{fig:bsd_compare_t5_sci5_barts_on_SARInBERTscore}
\end{figure}

\begin{figure}[]
  \centering
  \begin{subfigure}{0.49\textwidth}
    \includegraphics[width=\linewidth]{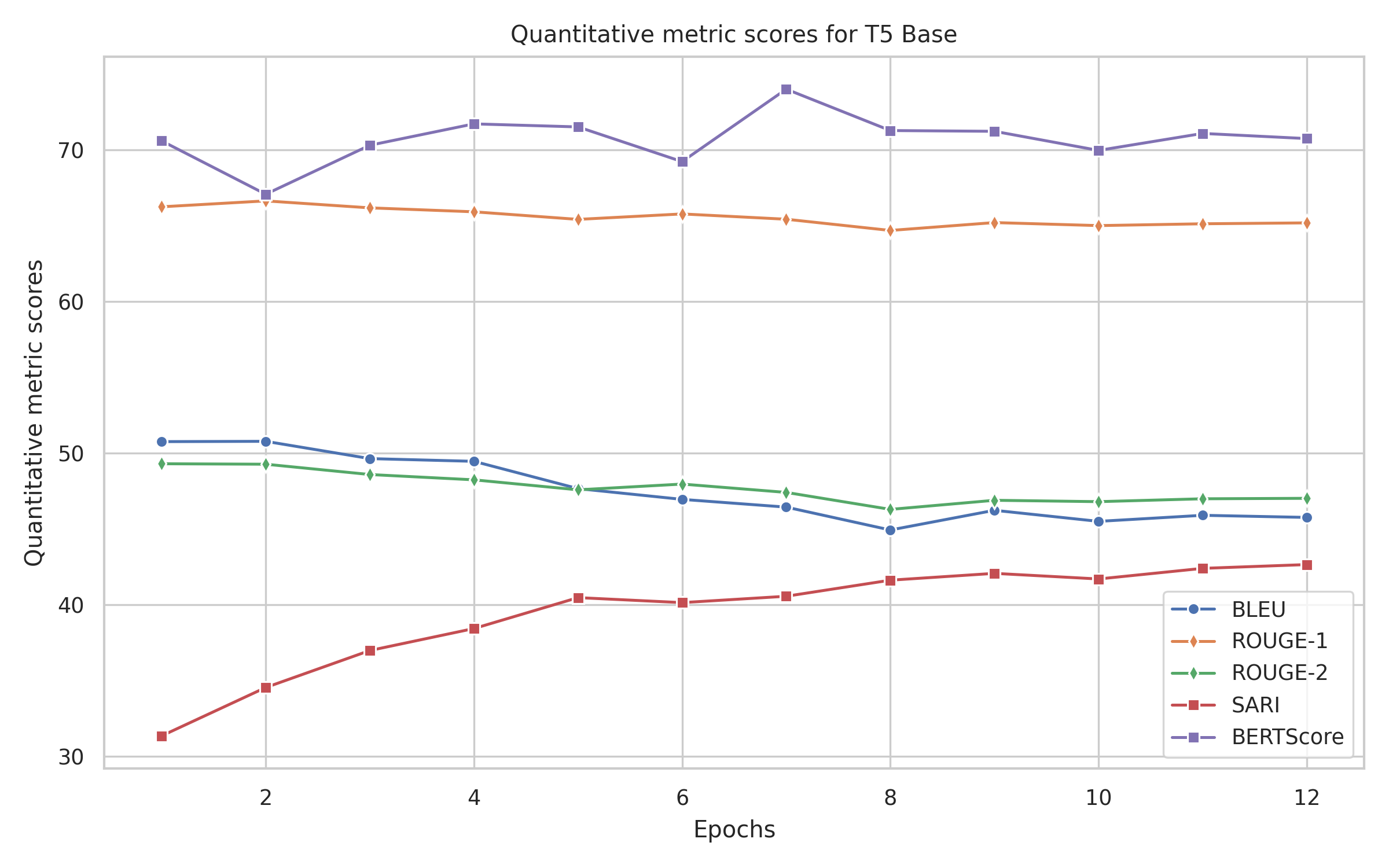}
  \end{subfigure}
  \hfill
  \begin{subfigure}{0.49\textwidth}
    \includegraphics[width=\linewidth]{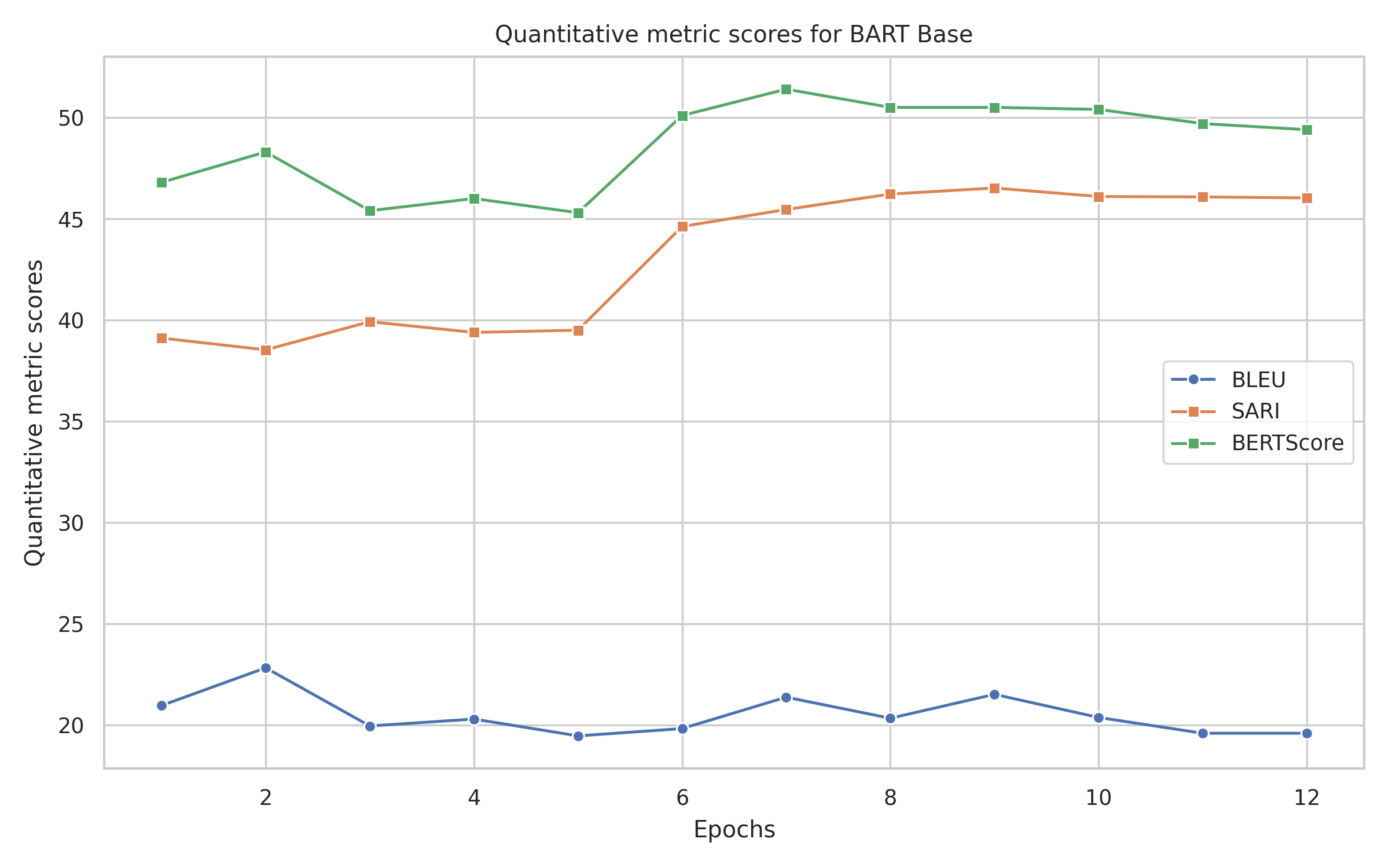}
  \end{subfigure}
  \caption{ Evaluation Scores of T5-base and BART-base on the Extracted Testing Set}
  \label{fig:t5base-BartBase-scores}
\end{figure}

To look into the details of model comparisons from different epochs on the extracted testing set, we present the learning curve of T5, SciFive, BART-base on WikiLarge, and BART-base on PLABA data in Figure \ref{fig:bsd_compare_t5_sci5_barts_on_SARInBERTscore}.
We also present the learning curve of T5 Base and BART Base using different metrics in Figure \ref{fig:t5base-BartBase-scores}.

\begin{table*}[t!]
  \centering
  \begin{tabular}{ |c||c|c|c|c|c|c|  }
      \hline
      Models & BLEU & ROUGE-1 & ROUGE-2 & ROUGE-L & SARI & BERTScore \\
      \Xhline{2\arrayrulewidth}
      T5 Small      & \textbf{49.86} & \textbf{65.94} & \textbf{48.60} & \textbf{63.94} & 33.38 & 69.58 \\
      T5 Base       & 43.92 & 64.36 & 46.07 & 61.63 & 44.10 & \textbf{72.62} \\
      T5 Large      & 43.52 & 64.27 & 46.01 & 61.53 & \textit{43.70} & 60.39 \\
      FLAN-T5 XL (LoRA) & 44.54 & 63.16 & 45.06 & 60.53 & 43.47 & 67.94 \\
      \hline
      SciFive Base  & 44.91 & 64.67 & 46.45 & 61.89 & 44.27 & 60.86 \\
      SciFive Large & 44.12 & 64.32 & 46.21 & 61.41 & \textit{44.38} & 72.59 \\ 
      \hline
      BART Base with CTs & 21.52 & 56.14 & 35.22 & 52.38 & 46.52 & 50.53 \\
      BART Large with CTs & 20.71 & 54.73 & 32.64 & 49.68 & \textbf{46.54} & 50.16 \\\hline
  \end{tabular}
  \caption{Automatic Evaluations of Encoder-Decoder Models on Extracted Testing Set. %T5, SciFive and BART Models with Control Token (CT) mechanisms on the Extracted Testing Set. FLAN-T5 XL used LoRA.
  }
  \label{table:evaluation_metrics_t5_sci5_bart_test}
\end{table*}

\begin{table*}[t!]
  \centering
  \begin{tabular}{ |c||c|c|c|c|c|c|  }
      \hline
      Models & BLEU & ROUGE-1 & ROUGE-2 & ROUGE-L & SARI & BERTScore \\
      \Xhline{2\arrayrulewidth}
      GPT-3.5       & 20.97 & 50.07 & 24.72 & 43.12 & 42.61 & 58.35 \\
      GPT-4         & 19.50 & 48.36 & 23.34 & 42.38 & 43.22 & 46.99 \\ 
      \hline
      BioGPT-Large (LoRA)  &  41.36 & 63.21 & 46.63 & 61.56 & 18.44 & 62.9 \\
      \hline
  \end{tabular}
  \caption{Quantitative Evaluations of GPTs and BioGPT-Large on the Extracted Testing Set}
  \label{table:evaluation_metrics_gpt_test}
\end{table*}

Because the fine-tuned T5-Base model has the highest BERTScore (72.62) and also a relatively higher SARI score (44.10), we chose it as one of the candidates for human evaluation. The other candidate is the fine-tuned BART Large with CT mechanisms which has the highest SARI score (46.54) among all evaluated models. Note that, SciFive Large has results close to the ones of T5 Base. In this case, we selected the smaller model for human evaluation.

\subsection{Human Evaluations}

In human evaluation, we randomly sampled 80 sentences from the test set split and evaluated the corresponding outputs of BART-large with CTs and T5-base with anonymisation setting. %anonymously. 
In the human evaluation form, we randomly assigned the order of the two systems' outputs and put them beside the input sentence. 
Based on the comparison of input and output sentences, the annotators need to answer two questions: ``To what extent do you agree the simplified sentence keeps the major information" and ``To what extent do you agree the simplified sentence is well simplified". 
The answer is limited to a 5-point Likert scale, from strongly disagree to strongly agree. 
%The sample form can be found in Table \ref{tab:Questionnaire} in the Appendix. 
There are 4 annotators in this human evaluation, one of the annotators is a native English speaker, and the others are fluent English users as a second language. 
Two annotators are final-year bachelor students, one annotator is a Masters candidate, and the last one is a Postdoctoral researcher. 
%All annotators hold a bachelor's or higher degree. 
Each annotator evaluated 40 sentence pairs with 50\% overlaps to make sure every sentence was evaluated twice by different annotators. The detailed human evaluation scores on the two selected models we evaluated are shown in Table \ref{tab: HumanEval_CToken_T5base} using the two designed criteria.

Based on the cross-annotation (overlaps), we calculated the inter-rater agreement level in Table \ref{tab:categories_CohenKappa} using Cohen's Kappa on cross models and Table \ref{tab:Krippendorff's_alpha_modelWise} using Krippendorff's Alpha with model-wise comparison. 
Both tables include the agreement levels on two sub-categories, namely ``meaning preservation" and ``text simplicity".
Because there is no overlap in the annotation tasks between annotators (0, 3) and annotators (1, 2), we listed the agreement levels between the available pairs of annotations, i.e. (0, 1), (0, 2), (1, 3), and (2, 3). 
The Cohen's Kappa agreement level presented in Table \ref{tab:categories_CohenKappa} shows the inter-rater agreement on the performance order of the two systems, whether one system is better than the other or tie. 
The Krippendorff's Alpha represented in Table \ref{tab:Krippendorff's_alpha_modelWise} shows the annotation reliability over all 5-Likert options. 

Based on the results from Table \ref{tab: HumanEval_CToken_T5base}, annotators show a different preference for the two systems. Despite the limited gap in the SARI score, BART with CTs shows a better capability to fulfil the simplification tasks. Yet regarding meaning preservation,  the fine-tuned T5-base performs better, as 
the BERTScore tells among the comparisons in Table \ref{table:evaluation_metrics_t5_sci5_bart_test}. % the gap and in meaning preservation.

From Table \ref{tab:categories_CohenKappa}, Annotators 0 and 1 have the highest agreement on ``meaning preservation'' (score 0.583), while Annotators 1 and 3 have the highest agreement on ``simplicity'' (score 0.238) evaluation across the two models. 
This also indicates that it is not easy to evaluate the system performances against each other regarding these two criteria. 

Model wise, Table \ref{tab:Krippendorff's_alpha_modelWise} further shows that Annotators 0 and 1 agree more on the judgements of fine-tuned T5 Base model on both two criteria of ``meaning preservation'' (score 0.449) and ``text simplicity'' (score 0.386), in comparison to the BART model.
On the fine-tuned BART Large model with CTs, these two annotators only agreed on the ``meaning preservation'' factor with a score of 0.441.
This phenomenon also applies to Annotators 0 and 2 regarding the judgement of these two models.
Interestingly, while Table \ref{tab:categories_CohenKappa} shows that Annotators 1 and 3 have a better agreement on ``simplicity'' judgement over ``meaning preservation'' using Cohen's Kappa, Table \ref{tab:Krippendorff's_alpha_modelWise} shows the opposite using Krippendorff's Alpha, i.e. it tells the agreement on ``meaning preservation'' of these two annotators instead.
This shows the difference between the two agreement and reliability measurement metrics.

\begin{table}[htb!]
    \centering
    \begin{tabular}{cp{2.5cm}c} \hline 
        Model & Meaning Preservation & Simplicity \\
        BART with CTs & 2.625 & 2.900 \\
        T5-base & 3.094 & 2.244 \\ \hline 
    \end{tabular}
    \caption{Human Evaluation Scores of Fine-tuned BART-Large with CTs and T5-Base Models on the Sampled Test Set}
    \label{tab: HumanEval_CToken_T5base}
\end{table}

\begin{table}[htb!]
    \centering
    \begin{tabular}{ccc}
        Annotator & Meaning preservation & Simplicity \\\hline
        0 \& 1 & 0.583 & 0.138 \\
        0 \& 2 & 0.238 & 0.126 \\ 
        1 \& 3 & 0.008 & 0.238 \\
        2 \& 3 & -0.130 & -0.014 \\\hline
    \end{tabular}
    \caption{Cohen Kappa among annotators over 3 categories ordinal - win, lose, and tie.}
    \label{tab:categories_CohenKappa}
\end{table}

\begin{table}[htb!]
    \centering 
    \begin{tabular}{p{1cm}p{2.1cm}p{1.8cm}p{1.2cm}}
        Anno. & Model & Meaning preservation & Simplicity \\ \hline 
        \multirow{2}{*}{0 \& 1} & T5-base & 0.449 & 0.386 \\
        & BART w CTs & 0.441 & 0.052\\
        \multirow{2}{*}{0 \& 2} & T5-base & 0.259 & 0.202 \\
        & BART w CTs & 0.200 & 0.007\\
        \multirow{2}{*}{1 \& 3} & T5-base & 0.307 & 0.065 \\
        & BART w CTs & -0.141 & -0.056\\
        \multirow{2}{*}{2 \& 3} & T5-base & -0.056 & 0.116 \\
        & BART w CTs & 0.065 & -0.285\\ \hline 
    \end{tabular}
    \caption{Krippendorff's alpha among annotators (Anno.) over the 5-Likert scale from strongly agree to strongly disagree.}
    \label{tab:Krippendorff's_alpha_modelWise}
\end{table}

\subsection{System Output Categorisation}

\begin{comment}
    using this doc \url{https://docs.google.com/document/d/1PVPL5PpcvfLactIZfktBxDnYilaQfYLwDSVE10MJvhY/edit?usp=sharing}
\end{comment}

We observe some interesting aspects of human evaluation findings by comparing the outputs of two models.
1) how to deal with the judgement on two models when one almost copied the full text from the source while the other did simplification but with introduced errors?
\begin{comment}
    For example, the source text ``
%T5 model kept almost the same as source: strong agree on meaning vs strong disagree on simplicity Vs CTs simplified output but changed some meaning.	
\textit{A national programme of neonatal screening for CAH would be justified, with reassessment after an agreed period}.'' is simplified by the BART-w-CTs model into \textit{``A national program of checking newborns for COVID-19 would be a good idea.''} However, ``CAH'' means ``Congenital adrenal hyperplasia'' instead of ``COVID-19''.
The T5-base model almost copied the same source text, producing \textit{``A national programme of newborn screening for CAH would be justified, with reassessment after an agreed period of time.''}
In this case, the T5-base model can get ``strongly agree'' (score 2) for meaning preservation but ``strongly disagree'' (score -2) for text simplicity, which would have an average score 0. 
BART-w-CTs can get ``strongly agree'' for text simplicity (score 2), but a lower score on meaning preservation, e.g. -2. In this case, the two models will be attributed the same score. Note that, for system selection, it might be a better choice to look into the separate dimensions of how the models perform. 
\end{comment}
2) Abbreviation caused interpretation inaccuracy. This can be a common issue in the PLABA task. For instance, the source sentence ``\textit{A total of 157 consecutive patients underwent TKA (n = 18) or UKA (n = 139)}.'' is simplified by the T5-base model into \textit{``A total of 157 consecutive patients underwent knee replacement or knee replacement.''} and by BART-w-CTs into \textit{``A total of 157 patients had either knee replacement or knee replacement surgery.'' }
Both these two models produced repeated phrases ``knee replacement or knee replacement'' due to a lack of meaningful understanding of ``TKA:  total knee arthroplasty'' and ``UKA: Unicompartmental knee arthroplasty''. A reasonable simplification here can be ``157 patients had knee surgery''.

\begin{figure*}[]
\centering
    \includegraphics[width=0.9\textwidth]{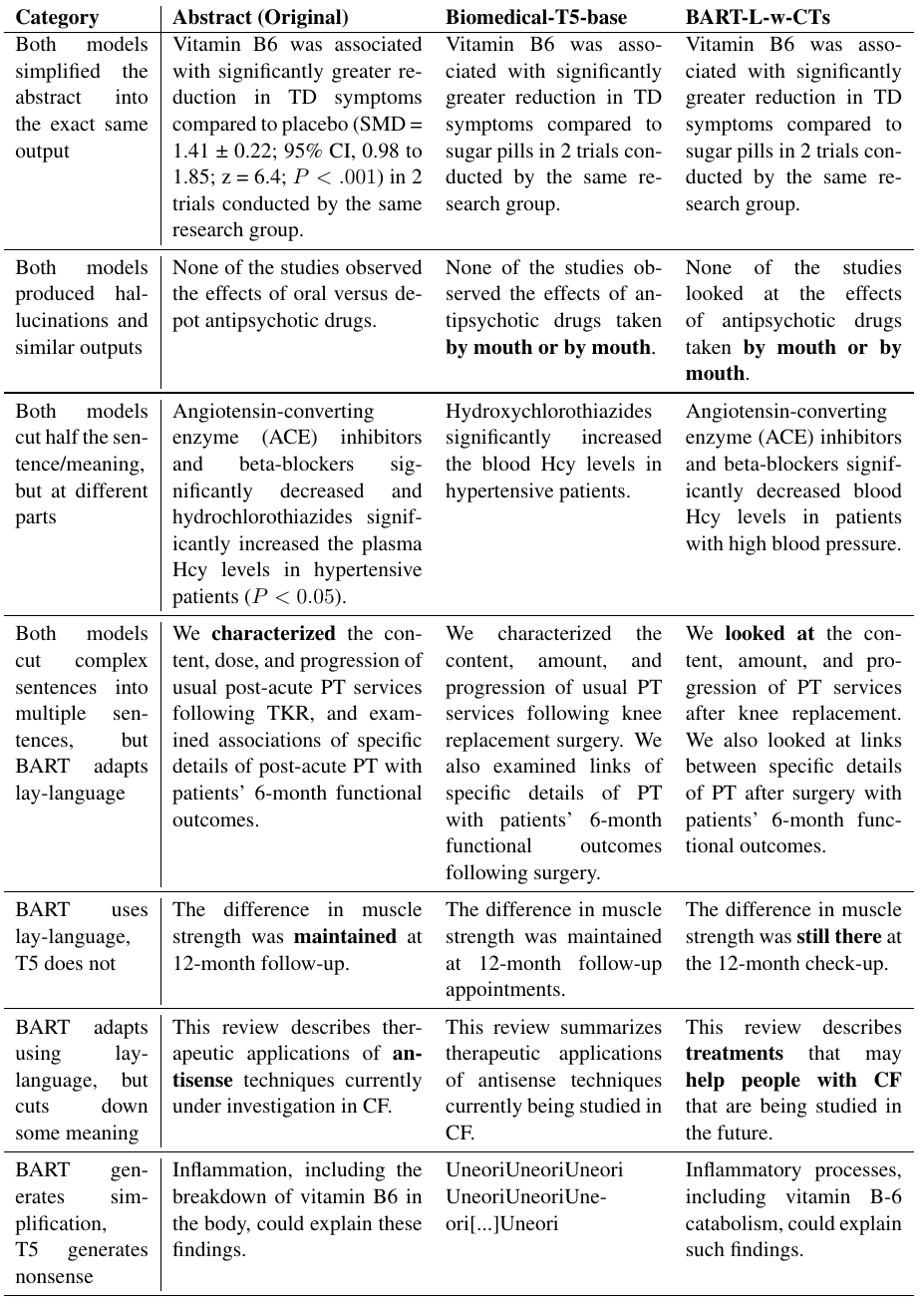}
    \caption{Segment-level examples of categorised outputs from two models.
    %(\textbf{do we put metrics used here?}).  %Full Training set: training+validation set;  %: Fine-tuned BART / Control Mechanisms %for PLABA2023 Challenge - to be replaced with this \url{https://drive.google.com/file/d/1GKHLtuR89PBTa0HGwm70V69dAnwrv-Q-/view?usp=sharing}
    %  One-Shot ChatGPT did not use the training and validation set we extracted. One-Shot: ChatGPT/GPT4; BioGPT another chunk. BioGPT trained on train+valid.  
    }
    \label{fig:segment_example_output}
\end{figure*}

We list more categories below by comparing outputs %and refer to Table \ref{tab:human_eval_catergorise} for examples 
from the two evaluated models. These categories apply to some segment-level comparisons, but \textit{not all the outputs}, e.g. in Figure \ref{fig:segment_example_output}.

\begin{itemize}
    \itemsep0em 
    \item Both models simplified the abstract into the exact same output
    \item Both models produced hallucinations and similar outputs
    \item Both models cut half the sentence/meaning, but at different parts
    \item Both models cut complex sentences into multiple sentences, but BART adapted lay-language
%    \item BART uses lay-language vs T5 does not
    \item BART adapted using lay language but cut out some meaning
    \item BART generated simplification vs T5 generated nonsense
    \item T5 did little simplification but maintained good meaning. %BART increases simplicity but loses some meaning
    \item T5 cut meaning; BART did not but maintained the same complexity as the abstract
    \item BART shifted the meaning
    
\end{itemize}

\section{Discussions}
Several aspects can be further improved or investigated: % from our work:
\begin{itemize}
    \item \textbf{On automatic metrics}. Based on the results depicted in Figure \ref{fig:t5base-BartBase-scores}, we acknowledge that SARI stands out as a more reliable metric than BLEU and ROUGE-1/2 for assessing the quality of generated simplifications. During the early training epochs, the model outputs closely resemble the input texts, which can lead SARI to assign lower scores compared to BLEU and ROUGE-1/2. This occurs because these metrics might be satisfied by the mirroring of n-grams between inputs and generated outputs during simplification. However, it is essential to acknowledge that, in the context of simplification, the generated output typically remains relatively close to the input. As a result, BLEU and ROUGE may exhibit consistent scores throughout epochs and may not effectively evaluate the quality of the generated texts. In contrast, BERTScore offers a different perspective by focusing on meaning preservation after simplification instead of simplification quality. If the generated outputs are copies of the input texts, BERTScore may still yield high scores although the model indeed performs poorly. Therefore, we use the combination of both metrics — SARI for evaluating generation quality and BERTScore for assessing meaning preservation — in order to select the best-performing models. It would be very useful to develop a new automatic metric that can effectively reflect both text simplicity and meaning preservation.
    \item \textbf{On human evaluations}. Due to the various backgrounds of annotators and the lack of proper training material and methods, it is difficult to build a standardised scheme to help annotators choose from the 5-Likert options. This means that the definition of ``Strongly Agree/Disagree'' may vary among annotators. In addition, there is no method to normalise the scores to a unified level in order to avoid the effects caused by differences in subjective judgements. Thus, we evaluate the two systems simultaneously and allow the annotator to decide on the performance order and performance gap. To reflect on the agreement of these two aspects, we calculated the inter-rater agreement in Cohen Kappa and Krippendorff's alpha. As shown in Table \ref{tab:categories_CohenKappa}, only annotators 0 and 1 show a decent agreement on meaning preservation, while the others show only limited agreement or even slight disagreement. Although human evaluation has long been the gold standard for evaluation tasks, there are few acknowledged works on a standard procedure to make it more explainable and comparable. In the future, a more standardised and unified process may be required.
    \item \textbf{On broader test sets}. In this work, we carried out an investigation using the PLABA data set. To be more fairly comparing selected models, we will include other related testing data on biomedical text simplification tasks. How to improve the number of references for the PLABA data set can be also explored.
 %:2bring-back?   \item  \textbf{On impacts of efficient training}. We applied LoRA efficient training for this work. Further investigation on the impact of LoRA regarding model prediction accuracy can be carried out. Alternative efficient training and fine-tuning methods can be also tested, in comparison to LoRA, on this specific task including Adapters and prefix-tuning \cite{li-liang-2021-prefix-tuning} and the methods applied by \cite{10.1145/3458754BLURB2021,Lan2020ALBERT:}.
\end{itemize}

\section{Conclusions and Future Work}
\label{sec_conclude}

%what do we conclude from our investigation? what can we do in the future based on what we did?

\begin{comment}
    Except for the situations that the two models both do a good job:
- BART-W-CTs are more effective on this task, which means it does something in most cases to simplify the text, but sometime it cut the meaning, and some times introduce new meaning / revise source.
- In comparison, T5 is less sensitive to this task; in many cases, it only copy the text or do minor changes.
\end{comment} 

%In this work, w
We have carried out an investigation into using LLMs and Control Mechanisms for the text simplification task on biomedical abstracts using the PLABA data set. Both automatic evaluations using a broad range of metrics and human evaluations were conducted to assess the system outputs.
As automatic evaluation results show, both T5 and BART with Control Tokens demonstrated high accuracy in generating simplified versions of biomedical abstracts. However, when we delve into human evaluations, it becomes clear that each model possesses its unique strengths and trade-offs.
T5 demonstrated strong performances at preserving the original abstracts' meaning, but sometimes at the cost of lacking simplification. By maintaining the core content and context of the input, it has proven to be over-conservative in some cases, resulting in outputs that very closely resemble the inputs therefore maintaining the abstract's complexity.
On the other hand, BART-w-CTs demonstrated strong simplification performances to produce better-simplified versions. However, it has shown a potential drawback in reducing the preservation of the original meaning. 

\begin{comment}
    This difference in approach is also reflected in the automatic metrics, where BART achieves the highest SARI score quantifying generation quality, but lags behind T5 in BERTScore, which measures meaning preservation.
In essence, while both models excel in generating simplified biomedical abstracts, T5 prioritises meaning preservation, while BART tends to favour more substantial simplifications. These distinctions are not only apparent in human evaluations but also supported by the differences in automatic metric scores.
\end{comment}

In future work, we plan to carry out investigations on more recent models including BioBART \cite{yuan-etal-2022-biobart}, try different prompting methods such as the work from \cite{cui-etal-2023-medtem2}, and design a more detailed human evaluation such as the work proposed by \cite{gladkoff-han-2022-hope} with error severity levels might shed some light on this.

%proposed \textit{Prefix-tuning} method
%Open Prompt for some models? or soft/mix prompts \cite{cui-etal-2023-medtem2}
%Human Evaluations with error severity? \cite{gladkoff-han-2022-hope}

%to try models: - BioBART \cite{yuan-etal-2022-biobart}?

%to try new/alternative efficient training/fine-tuning: BLURB, ALBERT?

\begin{comment} to look back:
    'This shows the difference between the two agreement and reliability measurement metrics.'
\end{comment}

%\section*{Acknowledgements} We thank the CSF3 server support from The University of Manchester. 
%We acknowledge the usage of the following packages: SKlearn, BERT, BioBERT, and ClinicalBERT.

%We are grateful for the support from the grant “Assembling the Data Jigsaw: Powering Robust Research on the Causes, Determinants and Outcomes of MSK Disease”. The project has been funded by the Nuffield Foundation, but the views expressed are those of the authors and not necessarily the Foundation. Visit www.nuffieldfoundation.org. We are also supported by the grant “Integrating hospital outpatient letters into the healthcare data space” (EP/V047949/1; funder: UKRI/EPSRC).

%\bibliographystyle{acl_natbib}
\bibliography{nodalida2023}

\bibliographystyle{IEEEtran}
%\bibliography{bibi}

%\begin{appendix} \end{appendix}

\end{document}